# A Comprehensive Overview of Recommender System and Sentiment Analysis

Sumaia Mohammed AL-Ghuribi [a,b] and Shahrul Azman Mohd Noah [a]
[a] Faculty of Information Science and Technology, Universiti Kebangsaan Malaysia, MALAYSIA
[b] Faculty of Applied Sciences, Department of Computer Science, Taiz University, YEMEN

**ABSTRACT**

Recommender system has been proven to be significantly crucial in many fields and is widely used by various domains. Most of the conventional recommender systems rely on the numeric rating given by a user to reflect his opinion about a consumed item; however, these ratings are not available in many domains. As a result, a new source of information represented by the user-generated reviews is incorporated in the recommendation process to compensate for the lack of these ratings. The reviews contain prosperous and numerous information related to the whole item or a specific feature that can be extracted using the sentiment analysis field. This paper gives a comprehensive overview to help researchers who aim to work with recommender system and sentiment analysis. It includes a background of the recommender system concept, including phases, approaches, and performance metrics used in recommender systems. Then, it discusses the sentiment analysis concept and highlights the main points in the sentiment analysis, including level, approaches, and focuses on aspect-based sentiment analysis.

## 1 INTRODUCTION

Recently, there has been a vast flow of information on the Web, and it continues to grow exponentially while providing users/customers with various resources about services such as products, hotels, and restaurants. Despite such data's benefits, the vast flow of information causes challenges for users to deal with and choose from a vast number of available options. This causes an information overload problem (Liu et al. 2011) and complicates the decision-making process. In this case, it is essential to filter the information to a limited amount based on the current user/customer preferences to assist them in making the correct decision (Hdioud et al. 2013). Such a filtering process is typically done by recommender systems (RSs), which are developed to solve the information overload problem by providing personalized suggestions of services (i.e. items) to specific customers according to their preferences (Adomavicius&Tuzhilin 2005; Ebadi&Krzyzak 2016).

Since RS emerged more than a few decades ago, the field has been of considerable importance in academia, business, and industry. They are widely used in various domains such as shopping (Amazon), music (Pandora), movies (Netflix), travel (TripAdvisor), restaurant (Yelp), people (Facebook), and articles (TED). With the recent advancement of e-commerce websites, it has been shown that RSs have a significant benefit in assisting users/customers to discover the relevant items that suit their needs and, most likely, to their preferences. For example, according to Amazon statistics in 2015, 35% of sales/revenues come from recommended items to users.

There are three main recommendation approaches which are content-based, collaborative-based, and hybrid. The content-based (CB) approach mines the appropriate recommendations for a user based on his recent behaviours according to what the user liked, bought, or watched (Pazzani&Billsus 2007). While the collaborative filtering approach (CF) generates the recommendation for a user based on the similarities among users who have similar preferences/interests to him in the past (Aciar et al. 2007). The last approach is the hybrid which integrates two or more recommendation components or algorithms implementations into a single recommendation system (Danilova&Ponomarev 2016). However, these classical RS approaches rely on a single-criterion rating (overall rating) as a primary source for the recommendation process. A single-criterion rating for a recommendation is insufficient to give an accurate recommendation because the overall ratings cannot express fine-grained analysis behind the users' behaviours. This made the RS a broad research topic that encouraged more research works to find practical solutions for improving the RS's performance.

At present, the sharing of experiences among customers is becoming a widespread phenomenon on social media sites. Many customers make decisions on consuming a service based on the opinions of others. Due to this phenomenon, there has been a rapid growth in the number of online opinions (i.e., user reviews). Each review expresses the customer's opinion of the used service, such as buying a product, watching a movie, or reserving a room. Such reviews are considered a valuable resource for both consumers and businesses. Despite these reviews' benefits, the extraction of useful information from such reviews is a huge challenge due to its large scale and distinct characteristics (Chen et al. 2015). Due to the reviews' characteristics, most RSs do not use them in generating recommendations because of the difficulties encountered by the machines to comprehend written natural language compared to other structured data sources (Musat et al. 2013).

Many fields are involved in processing textual reviews and extracting valuable information from the reviews, such as natural language processing, text mining, and opinion mining (or sentiment analysis). Sentiment analysis focuses on predicting the positive or negative polarity of the given entity.

Although there are some difficulties in processing users' reviews, there are significant advantages that RS can get benefit from them to enhance its performance. The following are some of the advantages of reviews that RSs can benefit (Al-Ghuribi&Noah 2019; Garcia Esparza et al. 2011; Jamroonsilp&Prompoon 2013; Zhang et al. 2015):

- Alleviate the data sparsity problem in the case of missing ratings. Reviews provide valuable and natural information about the user's interests which can be extracted and inferred.

- Relieve cold start problem either for a new user or a new item. Reviews can solve this type of problem by providing information that is used to improve recommendations (Wang et al. 2013).

- In the case of dense data, the reviews still provide valuable and detailed information that can be used to enhance the recommendation accuracy, such as: check the rating quality (compared both the user's star rating with the inferred rating from the review's text), derive users' aspects or context-dependent-aspect or context- independent-aspect preferences.

- Reviews help construct both the user and item profiles precisely because they contain much finer-grained sentiment trends for various features of a single item.

This paper presents a comprehensive background related to the recommender system and sentiment analysis. Firstly, it includes a background of the recommender system concept, including phases, approaches, and performance metrics used in recommender systems. Then, it discusses the sentiment analysis concept and highlights the main points in the sentiment analysis, including level, approaches, and focuses on aspect-based sentiment analysis.

The paper proceeds as follows: Section 2 presents an overview of the recommender system. Section 3 discusses an overview of the sentiment analysis. Finally, the paper ends with a conclusion presented in Section 4.

## 2 RECOMMENDER SYSTEM

In daily life, human depends on recommendations from others, either by word of mouth, movie and book reviews, letters of recommendation, restaurant and hotel guides or by some other means (Resnick&Varian 1997). The recommender system assists this natural social process. As a result, a comprehensive description of the recommender system is given in the following subsections, including its definition, phases, goals, approaches, different algorithms, and performance metrics.

## 2.1 Overview of the Recommender System

Nowadays, the recommender system (RS) is considered one of the most important digital world tools. RS success started in the past two decades due to information overload problems to serve as an information filtering system (Resnick et al. 1994). Generally, the term information filtering refers to two processes, (filtering in) aims to find the desired information for users, and (filtering out) aims to eliminate undesired information for users (Resnick et al. 1994). Since then, RSs have been incorporated and applied on several platforms, such as e-commerce and social networks (Lu et al. 2018). Besides, RS is a trending topic in the academic sector since several research works have been carried out and continued in the RS field. The problem of RSs can be identified as a way to assist users/customers in discovering relevant items to suit their needs and, most likely, to their preferences (Adomavicius&Tuzhilin 2005). There are many definitions of RSs, including:

- A **tool** to mine items and/or collect users' opinions to help users in their search process and suggest items related to their preferences (Hdioud et al. 2013; Kermany&Alizadeh 2017).

- A **program** or software for content filtering that attempts to reduce the information overload problem, where users encounter a flood of data on the Web, by recommending personalized items to users depending on the items' information and/or users' preferences (D'addio&Manzato 2015; Lakiotaki et al. 2011; Wang et al. 2018).

- A **system** to manage information overload problem by collecting information, guiding users in a personalized way, and providing individualized recommendations as output when there are many possible alternatives to choose from (Chen et al. 2015).

RS typically comprises three components: user/customer data (i.e., personal data such as interest, purchases history, ratings, and reviews), item data (i.e., item specifications and features), and the filtering techniques that use the two previously mentioned data to filter the items and recommend those that are close to the user interest. A general illustration of the RS architecture is depicted in Figure 1.

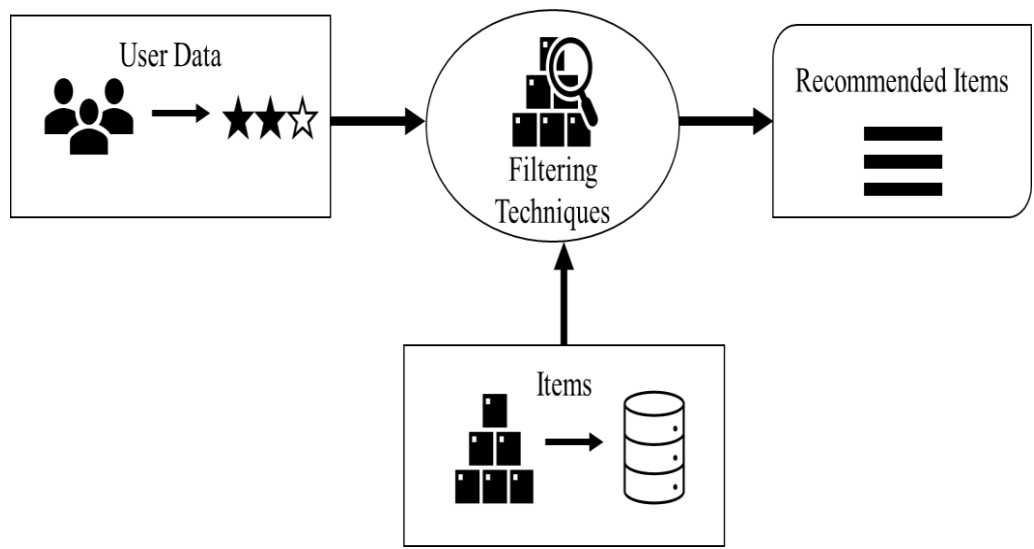

Figure 1    General RS architecture

Precisely, a RS model consists of two sets (Users, Items) and a utility function. All users are included in the Users set, and the Items set contains all the items that can be recommended to users. The utility function calculates the suitability of a recommendation to user $u \in$ *Users* an item $i \in$ *Items*, which is declared as R: *Users×Items*→$R_0$, where $R_0$ is equal to either a real number or a positive integer within a specific range (Adomavicius et al. 2011).

Typically, RS works through three phases (Adomavicius et al. 2011; Adomavicius et al. 2005; Ricci et al. 2011), as shown in Figure 2. The phases are as follows:

- **Modelling Phase**: This phase focuses on preparing the data that will be used in the subsequent two phases. There are three cases for that, the first is building a rating matrix that contains the users as records, items as attributes, and the value of each matrix's cell is the rating done by a user for a specific item. Second, building a user profile is mostly a vector for each user that explains his preferences of an item as a whole or on some aspects of the item. Third, building an item profile that contains the features of a specific item.
- **Prediction Phase**: This phase aims to predict the rating or score of unseen/unknown items for a specific user through a utility function depending on the extracted information during the modelling phase.
- **Recommendation Phase**: This phase is an extension of the prediction phase, where various approaches are applied to support the user's decision by filtering the most suitable items. It recommends/proposes new items to the user (i.e. a set of top-N items with the highest-predicted ratings) that is most likely attractive to him.

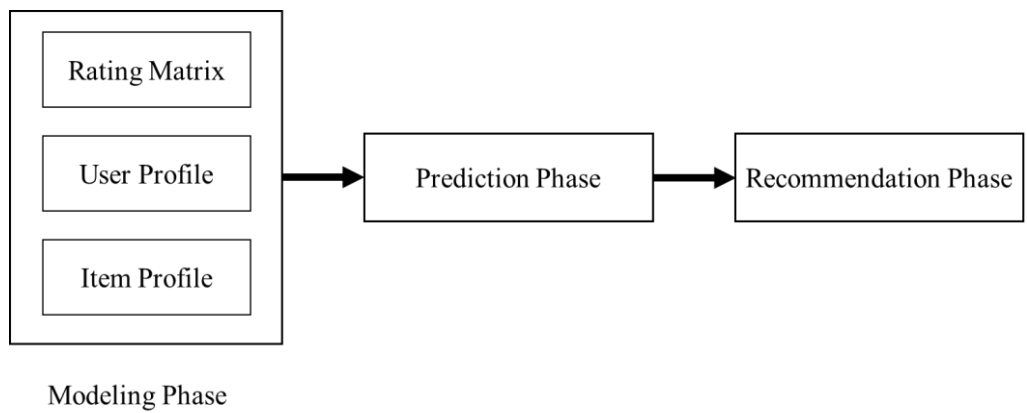

Figure 2    The phases of the recommender system

RS can benefit users, e-commerce websites, and many online companies. Currently, with the enormous spread of online shopping, many users rely on RSs to receive personalized recommendations and minimize the huge transactions for selecting items (Isinkaye et al. 2015). An example is shown in Figure 3, taken from the Amazon website to illustrate the recommendations for items during online shopping. Typically, users use this type of application to get a specific service, such as booking a hotel/ticket/restaurant, purchasing a product, watching a movie, etc. In this example, buying a face mask product is chosen as the service. It is obvious that several types of feedback are generated by users who consumed this item (i.e., purchasing this face mask) that convey their opinions on this product. The feedback is determined with red circles; users give either ratings for the whole product (i.e., the rating scale of ratings is from 1 to 5) or for a specific product feature (such as comfort, thickness, value of money, etc.).

Moreover, users may write reviews besides the ratings, such as those who are determined with brown rectangles, some of the users who write reviews provide images as well. It is very evident that these evaluations will influence the user's opinion and encourage him to buy or not to buy the product. In addition to the valuable feedback given by the Amazon application to assist users, it also provides users with recommendations based on some techniques. In this example, the given recommendations are specified in black circles, such as recommending products frequently purchased together with this face mask. Or recommending products with high ratings, or recommending products that have similar features to this face mask, or recommending products based on the user's browsing history.

Figure 3        Example of Amazon application recommendations for products

## 2.2 Goals of the Recommender System

The primary goal of the RSs is to solve the problem of information overload due to the exponential growth flow of information on the Web to provide users/customers with different sources about services such as products, hotels, and restaurants (Ebadi&Krzyzak 2016). To accomplish this primary goal, RS generally considers several common objectives, such as relevance/accuracy, novelty, serendipity, and diversity. These four objectives are briefly described as follows:

- **Relevance/ Accuracy -** Relevance and accuracy are synonymous expressed as the most prominent evaluation metrics for the RS. Users are more concerned with selecting items that are more relevant to their interests. As a result, RSs focus on recommending items to the user that are relevant to his preferences. The RSs accuracy is determined based on how much the item is relevant to user needs and liked by the user (Shi et al. 2014). The high values of the prediction performance metrics such as precision, recall, and F-measure reflect

    the RS model's good accuracy. On the other hand, lower values for the error metrics such as MAE and MSE indicates better RS accuracy.

- **Novelty -** Novelty is a fundamental aspect of the success of recommendations and an essential metric of customer satisfaction (Zhang 2013). RS novelty is the ability of an RS to generate novel recommendations for users. The recommendation is considered novel if it satisfies three characteristics (Zhang 2013): unknown, satisfactory and dissimilarity. **Unknown** refers to items that are not known to the user. **Satisfactory** relates to items that were satisfied by the user, whereas **dissimilarity** relates to items that are dissimilar to the items in the user profile. Novelty is contrary to popularity; the more popular the recommended items are for the user, the less novel the items are for the RS performance.

- **Serendipity -** The most closely related serendipity concept is unexpectedness, involving the user's positive emotional reaction to previously unknown serendipitous items (Adamopoulos&Tuzhilin 2014). Serendipitous items denote unexpected, novel, and relevant items to a user (Kotkov et al. 2017). RSs that recommend serendipitous items to users will significantly improve sales and create trust relationships with the users. Recommending items with fortunate discoveries will encourage the user's curiosity to enhance the user experience with the system in place.

- **Diversity -** Diversity is the opposite of similarity, RSs that provide diverse recommendations solve the over-fitting problem and increase the user's experience with the RS (Kunaver&Požrl 2017). Diversity generally applies to a set of items that have to do with how different the items are compared to each other (Castells et al. 2015). In other words, it relates to the internal differences within parts of an experience. Diversity will usually ensure that users are not dissatisfied by continually receiving the same recommendations at all times (Lu et al. 2015).

## 2.3 Recommender System Approaches

Approaches to the recommendation are usually categorized into three main categories: content-based, collaborative filtering, and hybrid. Figure 4 illustrates the three categories of the RS models.

### a. Content-Based Approach

A content-based (CB) approach mines the appropriate recommendations for a user based on his recent behaviours according to what the user liked, bought or watched (Pazzani&Billsus 1997). It generates the user profile from previously selected items by characterizing the user according to the item features and recommends items to the user based on the items with similar characteristics to the items that the user liked before (García-Cumbreras et al. 2013). It characterizes each user without having to compare his preferences to other users. Put differently, it does not use information about other users' preferences or the similarities with other users (Aciar et al. 2007; García-Cumbreras et al. 2013). A simple description of the CB approach is given in Figure 5. The process of the CB approach can be summarized into the following steps (Blanco-Fernandez et al. 2008; Cantador et al. 2008; Hdioud et al. 2013):

i. **Item representation**: The information source of the item description is used to extract the item's characteristics (i.e. features) to produce the structured item's representation.

ii. **Learning the user profile**: A user profile is generated from the previous user's behaviours (i.e. explicit and implicit feedback) such as like/dislike of an item; assign a score to an item (rating), or writing a textual opinion about an item (comment).

iii. **Recommendations' generation**: A list of items is recommended to the user by comparing the item's features with the users' profile. The most likely items to be attractive to the user are added to the list (i.e. top-ranked items).

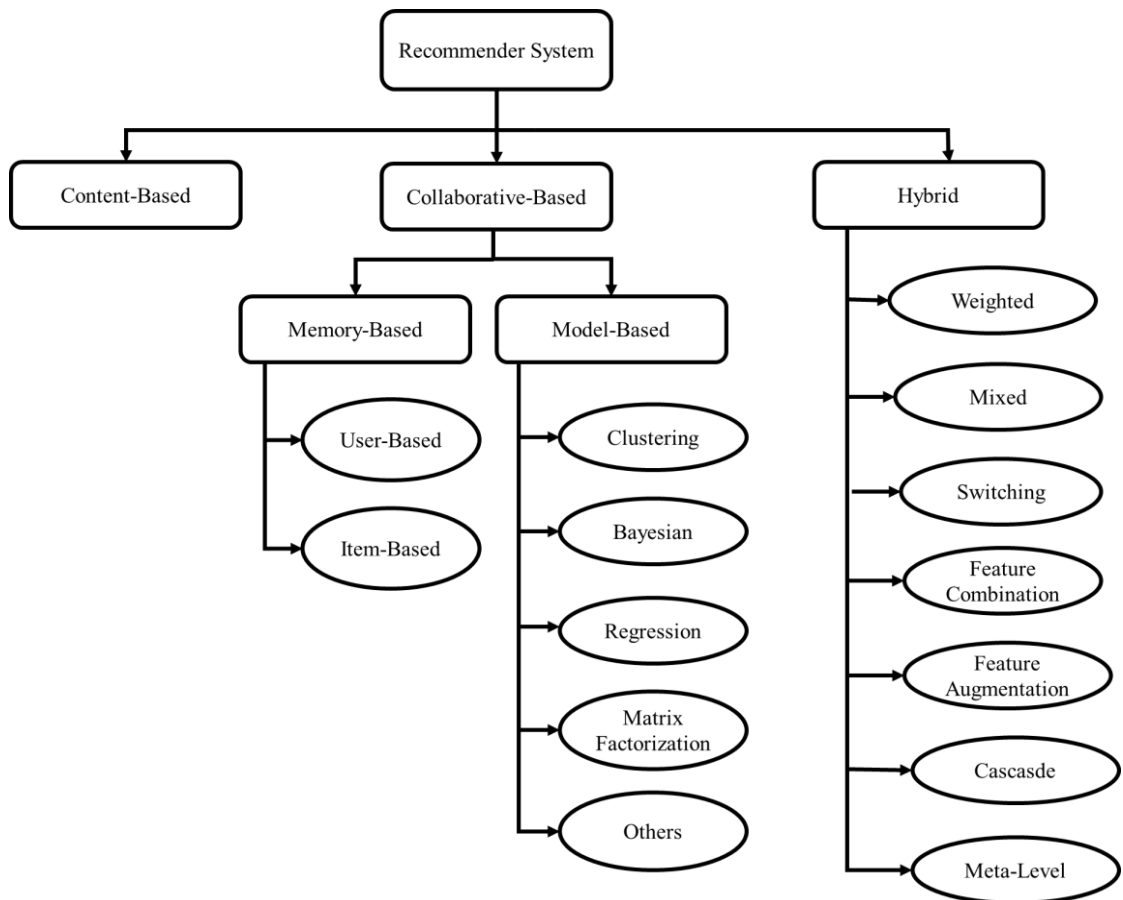

Figure 4     Recommendation system approaches

This type of approach has been implemented in many domains (Chen et al. 2015), especially in recommending items that contain textual information such as websites, news, and articles. It also recommends activities such as travel, tourism, e-commerce, and TVs (García-Cumbreras et al. 2013). This approach is preferred for moderate-sized items.

Some of the CB approach advantages are:

i. It can explain how to recommend specific items (i.e. present the logic behind their recommendations) by providing a list of content features. This, in turn, can strengthen the user's confidence about the RS that reflects his preferences (Aciar et al. 2007).

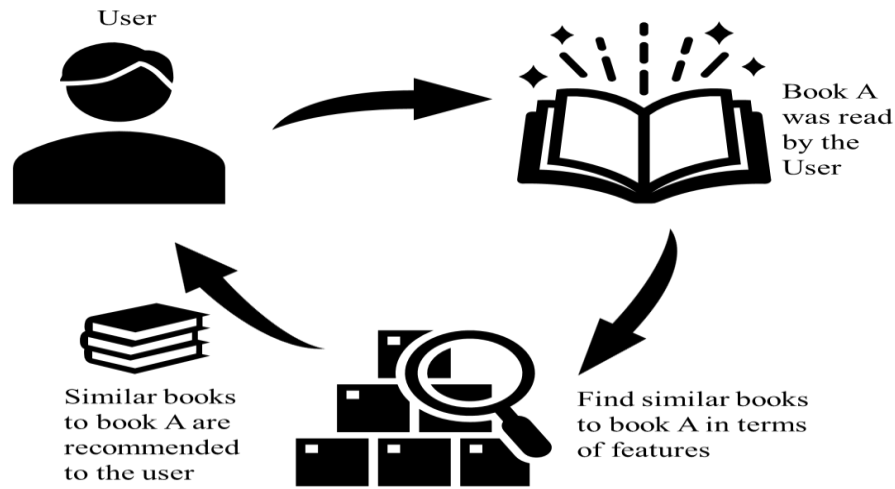

Figure 5    Simple illustration for CB approach

Some of the CB approach advantages are:

ii. Since this approach relies on the content of each item, not the ratings of other users, it gives several advantages as follows (Pazzani&Billsus 2007):
- It offers a high level of personalization in the recommendations.
- It is scalable in terms of the number of users.
- It can make recommendations for users with peculiar interests.
- It has high security from malicious item creation and allows users to prevent viral marketing.

On the other hand, CB approaches have some disadvantages, such as:

i. The vast size of the items is considered a major problem because when the recommendation is made, every item's content must be examined to discover items that are most likely relatable to the user's interest (Pazzani&Billsus 2007). This task is error-prone and time-consuming (D'addio et al. 2014).

ii. User profiles are built based on the static characteristics of the items. As a result, there is a high probability that different users have similar profiles even if they have various preferences among these items, just because they commented on the same items (Chen et al. 2015).

iii. The over-specialization problem occurs in this type of approach because users do not receive various or new items because of the restriction in their profile regarding the description of similar items (D'addio et al. 2014).

**b.    Collaborative-Based Approach**

The collaborative filtering approach (CF) is the most popular technique used in RSs (Yang et al. 2016). It generates the recommendation for a user based on the similarities among users who have similar preferences/interests to him in the past. This approach is based on the following hypothesis: people who agreed with a user in the past will also agree in the future (Aciar et al. 2007). A basic description of the CF approach for explaining the previous hypothesis is shown in Figure 6. CF identifies the new user-item association by determining the relationships between users and the interdependencies between items (Yang et al. 2016). It uses the implicit knowledge of a community of users on used items to identify those items' relationships to other users who have not used/seen those items within the community (García-Cumbreras et al. 2013). This can be represented as a user × items matrix in which each cell represents the user rating of a particular item.

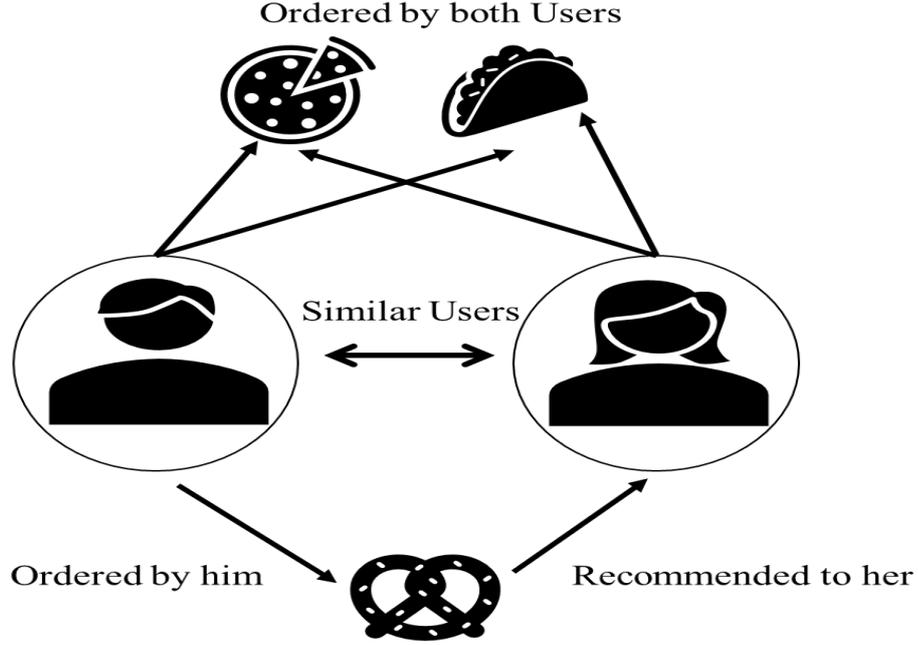

Figure 6        Simple illustration for CF approach

The first CF framework for RS was developed by Resnick et al. (1994) called GroupLens. It recommends articles to the Netnews clients using the rating server, named Better Bit Bureaus (BBB), which gathers users' rating to predict other user's scores on articles based on the CF hypothesis, clients who agreed to the rating of articles in the past they will probably agree in the future.

CF can be grouped into two classes memory-based and model-based (Chen et al. 2015). The memory-based CF type is a heuristic algorithm that predicts the item's rating based on other users' ratings. It can be classified into two methods (Adomavicius&Tuzhilin 2005): user-based and item-based; the former identifies a set of neighbours (i.e. like-minded users) for a target user using ratings then recommend a set of items that interest his neighbours. While the latter recommends items to a target user that are similar (i.e. has shared features) interests in the items that the user purchased, viewed or liked before. Figure 5 shows a comparison between user-based and item-to-item CF recommendations. In the user-based approach, one can see that users 'User 1' and 'User 3'are similar because they liked common items. Thus, 'User 3' is recommended to use the items that 'User 1' likes. Instead, with the item-to-item approach, similarities between items are calculated based on which users liked them. One can see that 'apple' and 'pear' are similar because they have been liked by users 'User 1' and 'User 2'. Thus, since 'User 3' has liked 'pear', then 'apple' will be recommended to him.

The Pearson correlation and cosine-based approach are the most frequently used to identify the user/item similarity (Liu et al. 2011). Follows is the details of the two approaches:

**i.    Pearson Correlation Coefficient**

Pearson correlation coefficient measures how strong a relationship is between two users/items using the following equation (Melville et al. 2002):

$$Sim(x,y) = \frac{\sum_{i=1}^{n}(r_{x,i}-\bar{r}_x)(r_{y,i}-\bar{r}_y)}{\sqrt{\sum_{i=1}^{n}(r_{x,i}-\bar{r}_x)^2}\sqrt{\sum_{i=1}^{n}(r_{y,i}-\bar{r}_y)^2}} \qquad (1)$$

Where Sim(x,y) in the above equation denotes the similarity between two users $x$ and $y$; $r_{x,i}$ is the rating given by user $x$ to item i; $\bar{r_x}$ is the average rating given by user $x$, while n is the total number of user-item space.

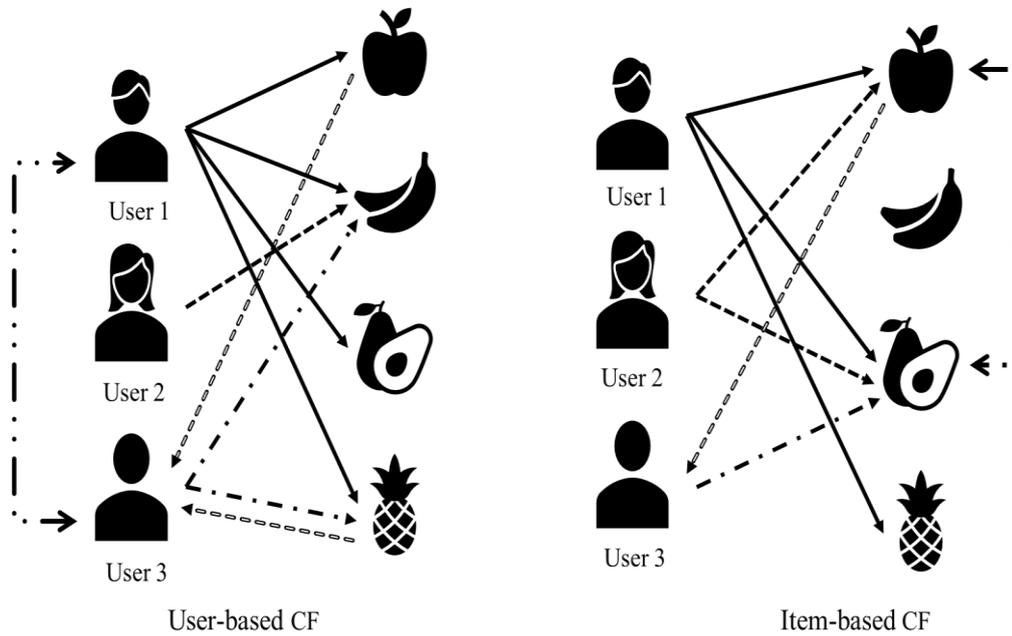

Figure 5  Simple illustrations for memory-based CF

### ii. Cosine Similarity

Cosine similarity is different from Pearson-based measure in that it is a linear algebra-based vector-space model rather than a statistical approach. It calculates the cosine angle projected in a multi-dimensional space between two vectors. The closer the value of the cosine to 1, the smaller the angle and the greater the similarity between vectors. This measure is widely used in information retrieval and texts mining to compare two documents (Isinkaye et al. 2015). The similarity of the two users $x$ and $y$ can be defined as:

$$Sim(\vec{x},\vec{y}) = \frac{\vec{x} \cdot \vec{y}}{|\vec{x}| * |\vec{y}|} = \frac{\sum_{i=1}^{n} r_{x,i} * r_{y,i}}{\sqrt{\sum_{i=1}^{n} r_{x,i}^2} \sqrt{\sum_{i=1}^{n} r_{y,i}^2}} \qquad (2)$$

As earlier mentioned that the user/item similarities are identified based on the similarity measures. For more details, users' similarity in the user-based is measured by comparing the ratings on the common items they rated. The item-based, on the other hand, measures the similarity between items, not users. It retrieves all items rated by an active user from the user-item matrix, decides how similar the retrieved items are to the target item, and then selects the *k* items that are the most similar to the target item. After the similarities are calculated, the prediction of the target item for the user is calculated using different prediction algorithms in order to recommend/not recommend this item to the user based on the predicted value. In section 2.4, the prediction algorithms will be addressed.

On the other hand, the model-based CF type (Su&Khoshgoftaar 2009) predicts the user's rating of unseen items by developing models using different representative techniques such as the clustering models, Bayesian networks, matrix factorization, and Markov decision process. A survey by Su and Khoshgoftaar (2009) provided a comparison between the CF classes, as shown in Table 1.

Table 1    Comparison between CF classes

| CF Class | Advantages | Shortcomings |
| --- | --- | --- |
| **Memory-based** | ▪ Easy to implement.<br>▪ Easy to add new data incrementally.<br>▪ The content of recommended items need not consider.<br>▪ Co-rated items have been scaled well. | ▪ Dependent on human ratings.<br>▪ In sparse data, the recommendation performance is decreased.<br>▪ New items/users cannot be recommended (i.e. cold start problem).<br>▪ The large datasets have limited the scalability. |
| **Model-based** | ▪ The scalability and sparsity problems are better addressed.<br>▪ The prediction performance is improved.<br>▪ An intuitive rationale is given for recommendations. | ▪ Building the model is expensive<br>▪ A trade-off is done between scalability and prediction performance.<br>▪ Useful information is lost through dimensionality reduction techniques. |

CF approaches possess many advantages compared to other approaches; some of the main advantages are:

- Serendipity where novel and unfamiliar items are recommended.
- Able to recommend more subtle items and can capture more nuances around items.
- Flexible and suitable for various domains.
- No need to analyse the items' contents.

Generally, the CF approach's performance depends on the availability of sufficient user participation (Aciar et al. 2007). It performs satisfactorily only when there is adequate rating information (Su&Khoshgoftaar 2009). Depending on the ratings exposed CF approach to the following issues (Yang et al. 2016), (Chen et al. 2011):

i. **Sparsity Problem**: One of the major problems that complicate the personalized item ranking process is data sparsity because items cannot be reliably linked to users (Musat et al. 2013), causing a limitation in the recommendation's effectiveness and limited coverage of recommendation space (Esparza et al. 2011). This problem occurs due to the following issues (Esparza et al. 2011; Leung et al. 2006):

    - Insufficient or missing information of either the user or item or both in the dataset during the process of filling the ratings (user-item) matrix. The complexity of gathering the items' ratings.
    - Expressing user's preferences about items as a rating is a complicated process.

ii. **Cold Start Problem**: This problem happens in the case of new users who do not provide any ratings yet or new items that have not been rated (Schein et al. 2002). It can be considered a particular case of the sparsity problem in which most of the cells of the item-user interaction matrix contain null values (Huang et al. 2004). The CF approach cannot generate accurate recommendations for new users or items without sufficient existing data (Chen et al. 2011).

iii. **Scalability**: The number of users and items in a system grows rapidly. For example, such a user's behaviour per day may result in his stored data reaching the size of TBs in some popular websites (Xin 2015). Furthermore, the RS should respond in less than a second to keep users satisfied and enable them to continuously engage with the RS (Xin 2015). As a result, both large-scale datasets and responding time create a challenge in designing efficient RS, and as a result, it demands colossal computing resources.

iv. **Rating bias**: In the CF approach, recommendations are based on users' ratings, but these ratings cannot show users' preferences or their clear opinion on some criteria, which makes it difficult in interpreting these ratings.

c. **Hybrid Approach**

This approach aims to mitigate the weakness of both CF and CB and benefit from their strengths by integrating two or more recommendation components or algorithm's implementations in a single recommendation system to enhance RS accuracy and gain better performance (Danilova&Ponomarev 2016; Hdioud et al. 2013). A simple illustration of the hybrid approach is given in Figure 6. When the hybrid approach is generated by hybridizing two or more algorithms, two major points must be taken into account (Hdioud et al. 2013): the first is the recommendation models that declare the required inputs and the determination on which the hybrid recommender will be based on. The second point is determining the strategy that will be used within the hybrid recommender (Burke 2007).

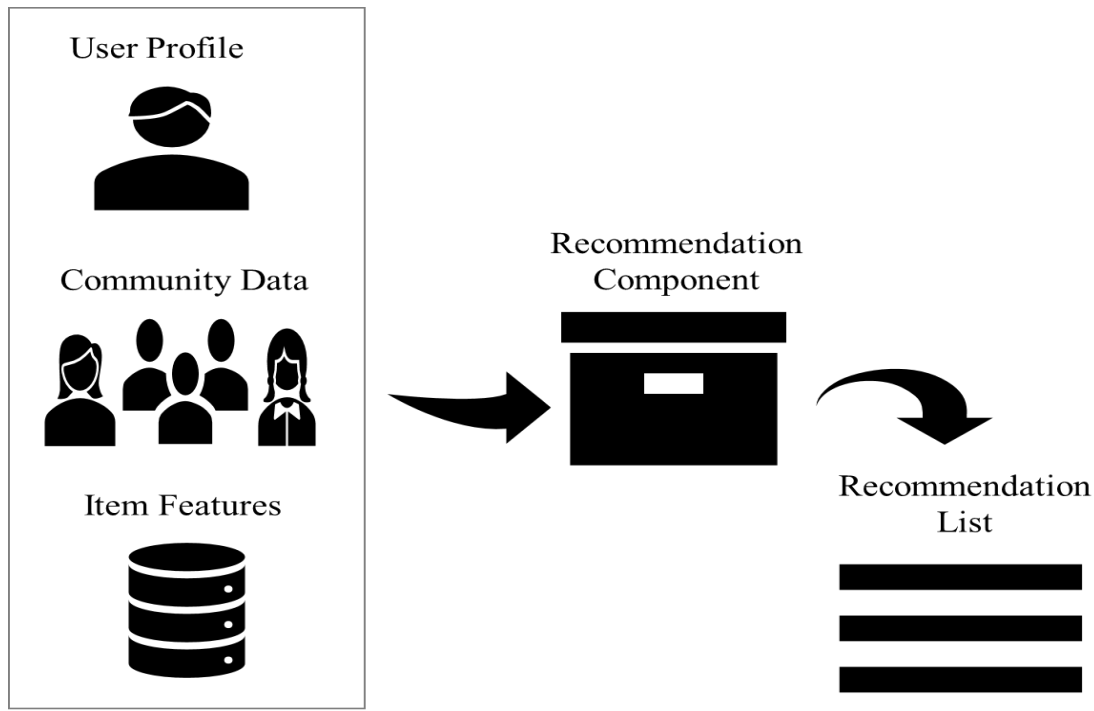

Figure 6 Simple illustrations for the hybrid approach

Seven different strategies are identified in hybridization (Burke 2007) as follow:

i. **Weighted** - Different recommendation algorithms are implemented, and their score (i.e. prediction) are combined numerically.
ii. **Mixed** - Recommendations from various recommenders are incorporated together at the same time.
iii. **Switching** - Two recommender systems that run on the same object are interchanged by the hybrid system based on a specific switching criterion.
iv. **Feature Combination** - Derived features from various sources of many recommender systems are joined into one recommender system.
v. **Feature Augmentation** - The output of the first recommender system will be used as an input to the second recommender system.
vi. **Cascade** - One recommender system has a higher priority than others and can refine the results given by a lower priority recommender system.
vii. **Meta-level** - The first recommender system is applied, and a sort of model is produced and used as an input to the next recommender.

Although hybrid approaches may overcome the limitation of both CB and CF approaches and enhance the prediction performance, it is expensive to implement, increases the complexity and needs external information that is mostly unavailable (Su&Khoshgoftaar 2009).

## 2.4 Prediction Algorithms

As earlier mentioned, the second phase of the RS phase is the prediction phase, which attempts to predict the rating/score of unseen/unknown items for a particular consumer. The prediction algorithms try to guess the rating a user is going to provide for an item; it is possible to define the problem of rating prediction as follows (Jahrer et al. 2010):

Suppose $R = [\![r_{u,i}]\!]$ is the U x M rating matrix: the U records are horizontal that represent the number of users, and the M attributes are vertical, representing the number of items. Each element of a matrix is denoted by a numerical value that reflects the ratings of user $u$ for item $i$. This matrix is usually sparsely filled since not all the items are commonly rated by a user. The purpose of the prediction model is to accurately predict missing values of this matrix, i.e., to generate $r_{u,i}$ prediction for how user $u$ will score an item $i$. An example of the rating matrix is shown in Figure 7; those items with the value '**?**' are the ones that the prediction algorithms try to predict their ratings.

The prediction algorithms use the historical record of ratings, item-related content, and user-related content to provide predictions (Papagelis&Plexousakis 2005). Many algorithms aim to predict user ratings. The two most valuable algorithms based on the Netflix Prize challenge are *k*-nearest neighbour (*k*NN) methods and the methods based on matrix factorization (Jahrer et al. 2010). Thus the following subsections describe these two algorithms.

Figure 7    Example of the rating matrix

### a.    *k*-Nearest Neighbour

The *k*-nearest neighbour (*k*NN) has emerged among the most common underlying algorithms for memory-based CF recommendation systems. The broad success it has achieved is due to the fact that it automates the process of gathering and merging content from human decisions (Lathia et al. 2008). By doing that, it can compute recommendations that implicitly have meanings, not just the closeness of the specifications between the two items (Lathia et al. 2008). The *k*NN was first developed by Evelyn Fix and Joseph Hodges in 1951 and later extended by Thomas Cover as a non-parametric, lazy learning method. It is used for regression and classification (Al-Ghuribi&Alshomrani 2013). The *k*NN algorithm assumes that similar things converge. In other words, similar things are close to each other, as in the famous quote, "Birds of a feather flock together".

The *k*NN makes no assumptions about the underlying distribution of data but relies on the similarity of item features (sometimes called proximity, distance, or closeness). The *k*NN assumes that the data is in a feature space since the points are in the feature's space; they have a distance notion. It also implies that each of the training data consists of a set of vectors associated with each vector and a class label. A single "*k*" number is provided; this number determines how many neighbours affect the classification.

The *k*NN is used as a prediction algorithm for the RS, and there are several variants of *k*NN approaches such as Basic *k*NN, *k*NN taking into account each user's mean ratings, and *k*NN taking into account each user's z-score normalisation. Table 2 shows the formula for each variant of the *k*NN algorithm. To clarify, a prediction rating $r_{u,i}$ of user $u$ on item $i$ in *k*NN using user's mean ratings is obtained by firstly selects the *k* best correlated (similar) users to user $u$. Then, the rating prediction for item $i$ is made from the weighted combination of ratings of the selected neighbours, which is measured as the weighted deviation from the mean of the neighbours (i.e., according to the degree of similarity shared with user $u$, the weight of each neighbour is determined) (Isinkaye et al. 2015).

Table 2      Descriptions for *k*NN algorithms

| *k*-NN Algorithm | Formula | | Details |
|---|---|---|---|
| Basic*k*NN | $p(u,i) = \dfrac{\sum_{x=1}^{n} sim(u,x) \cdot r_{x,i}}{\sum_{x=1}^{n} sim(u,x)}$ | (3) | • $\mathbf{p(u,i)}$: prediction function of user u on item i |
| *k*NNWithMean | $p(u,i) = \overline{r_u} + \dfrac{\sum_{x=1}^{n} sim(u,x) \cdot (r_{x,i} - \overline{r_x})}{\sum_{x=1}^{n} sim(u,x)}$ | (4) | • $\mathbf{sim(u,x)}$: similarity value between the two users u and x <br>• $r_{x,i}$: rating done by user $x$ for item $i$ <br>• $\overline{r_u}$ : mean ratings of user $u$ |
| *k*NNWithZScore | $p(u,i) = \overline{r_u} + \sigma_u \dfrac{\sum_{x=1}^{n} sim(u,x) \cdot (r_{x,i} - \overline{r_x})/\sigma_x}{\sum_{x=1}^{n} sim(u,x)}$ | (5) | • $\sigma_u$ : z-score normalization of user $u$ <br>• $n$: size of *k*-nearest neighbours |

**b.**     **Matrix Factorization**

Matrix factorization (MF) is one of the popular algorithms of the model-based CF that has obtained excellent results in the Netflix prize problem (Koren et al. 2009). It attempts to compact large databases into a single model and apply a reference mechanism to perform recommendation tasks. The MF algorithm is explicitly used to depict items and users using vectors of latent factors inferred from items' ratings (Isinkaye et al. 2015). Basically, MF tries to find a low-rank approximation to the rating matrix (Chen&Peng 2018). The fundamental assumption of the MF model is that it is possible to embed user preferences and item characteristics in the same subspace. If the latent factor vector of the user matches the latent factor vector of the item efficiently in the subspace, it is likely that the user is interested in the item (Chen&Peng 2018).

MF maps users and items to a shared dimensionality subspace, based on the assumption that the rating matrix has a low rank, such that a user-item rating can be modelled as the user's inner products of the item and user latent factor vectors in that subspace (Chen&Peng 2018). A simple illustration for the MF concept is depicted in Figure 8, in which the rating matrix R (M x N) decomposed into two smaller matrices, P (M x K) and Q (K x N), where K is the number of features.

Several models are built using MF; Singular value decomposition (SVD) is the most traditional one used for rating prediction (Alter et al. 2000). It has gained popularity due to its good accuracy and scalability (Chen&Peng 2018). Then an extension of the SVD model is proposed by Koren (2008) that incorporates implicit feedback to enhance prediction accuracy. Furthermore, there is another MF model proposed by Lee (1999) that factorized the rating matrix into two matrices with the property that there are no negative elements in all three matrices; this model called the Non-negative matrix factorization (NMF).

Typically, the prediction rating using the MF model is given by the inner product of a user feature vector and the item feature vector (i.e., each item $i$ is associated with an item-factors vector, and each user $u$ is associated with a user-factors vector). For more details, in the SVD model, two-factor matrices are learned through stochastic gradient descent. The first is the user factors matrix P = [$p_1, p_2, p_3... p_M$], and the second is the item factor matrix Q = [$q_1, q_2, q_3... q_K$]. The prediction rating of user $u$ on item $i$ is done by taking an inner product of a user feature vector $p_m$ and the item feature vector $q_k$ as the following equation: $\hat{r}_{ui} = p_u^T q_i = \sum_{i=1}^{k} p_{u,k}\, q_{k,i}$ .

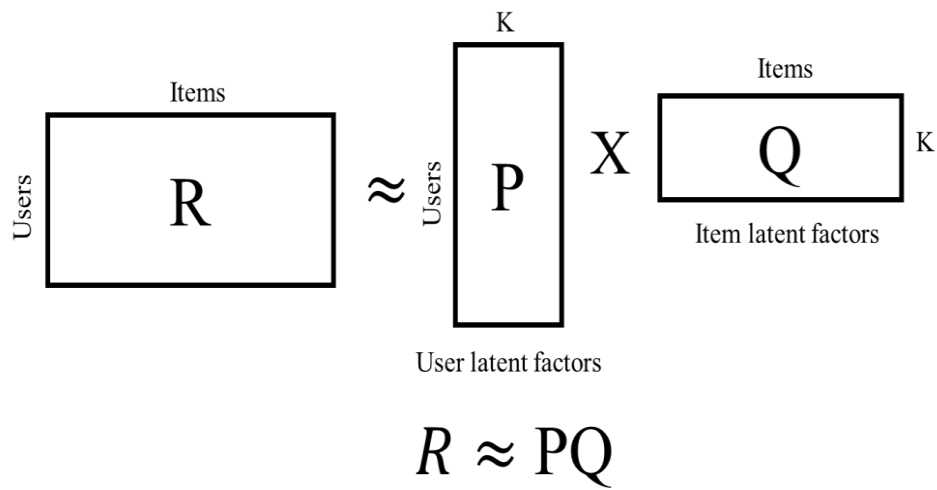

Figure 8      Illustration for the matrix factorization idea

## 2.5 Evaluation Metrics for the Recommender System

As discussed in section 2.2, RSs have different goals that vary from the main one to the sub-goals. A particular RS's success and efficiency often depend on the RS's primary objective and the domain characteristics to which it is applied (Schröder et al. 2011). Evaluation metrics for RSs can be broadly categorized into three (Herlocker et al. 2004): predictive accuracy metrics, classification accuracy metrics, and rank accuracy metrics.

### a. Predictive Accuracy Metrics

Predictive accuracy or rating prediction metrics answer the question of how similar a recommender's predicted ratings are to actual user ratings (Schröder et al. 2011). These metrics evaluate the RS performance by measuring the average error between the real ratings and the ratings predicted by the system. These metrics are mostly used as they are easy to compute and understand and particularly useful for evaluating tasks in which the predicting rating will be displayed to the user (Herlocker et al. 2004).

As discussed in the works (Gunawardana&Shani 2009; Herlocker et al. 2004), there are three crucial prediction accuracy metrics used: Mean Absolute Error (MAE), Mean Square Error (MSE) and Root Mean Square Error (RMSE). Usually, these metrics calculate the error difference between the predicted ratings and the user's actual ratings; thus, a lower value implies a better RS performance for these metrics. Table 3 defines the formulas of the three metrics.

### b. Classification Accuracy Metrics

Classification metrics determine the extents to which a RS makes correct or incorrect decisions about whether an item is good (i.e., good means relevant to user preferences) (Herlocker et al. 2004; Musat et al. 2013). Generally, a list of items for users is recommended by the RS. In general, they are organised horizontally or vertically, and users only care about the front parts of most items or the back parts of the items. A top-n recommendation is the name of this way of recommendations. The three most popular metrics are used to evaluate the RS recommendations' efficiency: precision, recall, and F-Measure.

Table 3    Predictive accuracy metrics

| Evaluation Metric | Formula | | Details |
|---|---|---|---|
| MAE | $MAE = \dfrac{\sum_{i=1}^{N}(p_i - r_i)}{N}$ | (6) | |
| MSE | $MSE = \dfrac{\sum_{i=1}^{N}(p_i - r_i)^2}{N}$ | (7) | ▪ $N$: is the size of the test set<br>▪ $p_i$: predicted rating calculated by RS<br>▪ $r_i$: actual rating given by the user |
| RMSE | $RMSE = \sqrt{\dfrac{\sum_{i=1}^{N}(p_i - r_i)^2}{N}}$ | (8) | |

### c. Classification Accuracy Metrics

Classification metrics determine the extents to which a RS makes correct or incorrect decisions about whether an item is good (i.e., good means relevant to user preferences) (Herlocker et al. 2004; Musat et al. 2013). Generally, a list of items for users is recommended by the RS. In general, they are organised horizontally or vertically, and users only care about the front parts of most items or the back parts of the items. A top-n recommendation is the name of this way of recommendations. The three most popular metrics are used to evaluate the RS recommendations' efficiency: precision, recall, and F-Measure.

The item set must be divided into two classes to calculate the previous metrics for RS (Gunawardana&Shani 2009; Herlocker et al. 2004): the first class determines the relevance of the item: relevant and irrelevant. The second class determines the quality of the returned items to the user selected/recommended and not selected/not recommended. The rating scale in specific datasets is not binary, so it needs to be converted into a binary scale. For example, the Amazon dataset's rating scale ranges from 1 to 5 to transform it to a binary scale; all the ratings of 1 to 3 are converted to *irrelevant,* and all the 4 and 5 ratings are converted to *relevant.* The item set classes are organized in Table 4 to be used later in the metrics' formula.

Table 4    Classes of item set

|  | **Selected** | **Not Selected** | **Total** |
|---|---|---|---|
| **Relevant** | $N_{rs}$ | $N_{rn}$ | $N_r$ |
| **Irrelevant** | $N_{is}$ | $N_{in}$ | $N_i$ |
| **Total** | $N_s$ | $N_n$ | $N$ |

Based on the item set classes shown in Table 4, the description of each of the three famous classification accuracy metrics is shown in Table 5.

Table 5    Classification accuracy metrics

| Evaluation Metric | Definition | Formula | |
|---|---|---|---|
| **Precision** | The ratio of relevant items selected to the number of items selected | $P = \dfrac{N_{rs}}{N_s}$ | (9) |
| **Recall** | The ratio of relevant items selected to total number of relevant items | $R = \dfrac{N_{rs}}{N_r}$ | (9) |
| **F-measure** | The harmonic mean of Precision and Recall that offers a better measure of the incorrectly classified cases | $F - score = \dfrac{2PR}{P + R}$ | (10) |

The formula shown in Table 5 is the standard one for the classification accuracy metrics; some works such as D'addio et al. (2017) do not measure these metrics for all the relevant items, a small k sample of the total ranking recommendations is only used as follows: precision@k, recall@k, and F-measure@k.

### d.    Rank Accuracy Metrics

Rank accuracy metrics measure a recommendation algorithm's ability to create a recommended order of items that matches how the same items would have been ordered by the user (Herlocker et al. 2004). Unlike the two previous accuracy metrics, this type does not seek to calculate the RS ability to accurately predict a single item's rating. Rank accuracy metrics are suitable for the RSs that provide the user with ranked recommendation lists and concern with differentiating between the "best" alternatives and just "good" alternatives of the recommended items. This type is suitable for the domains interested in ordering items and emphasizes the differences between the elements (Herlocker et al. 2004).

The most famous examples of the metrics in this category include Hit Ratio and Mean Reciprocal Rank (Chen&Chen 2014). Hit Ratio at N (HR@n) is a metric for calculating how many "hits" are done in an n-sized list of ranked items. A "hit" can describe an activity done by the user, such as liking, purchasing, clicking on, or saving/favourite. HR@n measures the percentage of RS successes by calculating the average probability that the pushed item will be recommended by a top N recommender (Sarwar et al. 2001). Simply, HR@n shows whether the desired items appear on the recommendation lists (presence) and how many times they appear (frequency) (Dias&Fonseca 2013). In contrast, Mean Reciprocal Rank (MRR) is a metric using to evaluate the generated recommendation lists in terms of determining the ranking position of the target user's choice in the recommendation list (Chen&Chen 2014). The Reciprocal Rank (RR) calculates the reciprocal of the rank at which the first relevant item was retrieved. If the relevant item was retrieved at rank 1, the RR is 1 if it was not

0.5 if the relevant document was retrieved at rank 2 and so on. When averaged across multiple queries, the measure is called the Mean Reciprocal Rank (MRR).

# 3 SENTIMENT ANALYSIS

Opinions (i.e., sentiments, emotions, attitudes) are a primary influence of our behaviours and fundamental to nearly all human activities. To a considerable degree, human views and interpretations of reality and their decisions are based on how others see and evaluate this universe (Liu 2012; Schouten&Frasincar 2016). As a result, people always seek out others' opinions when they need to make a decision; this is true for people and organisations. Opinions are the topics of the sentiment analysis field. This field has evolved to be one of the most active research areas in data mining, text mining and natural language processing since the beginning of 2000 due to the availability of a massive amount of opinionated data published in digital formats (Liu 2012). As a result, a comprehensive description of the sentiment analysis is given in the following subsections, including its definition, levels, tasks, approaches, and aspect-based sentiment analysis.

## 3.1 Overview of the Sentiment Analysis

Sentiment Analysis (SA) or Opinion Mining (OM) are two interchangeable terminologies to a discipline derived from artificial intelligence, information retrieval, and natural language processing (Schouten&Frasincar 2016). SA is the research area that analyses people's sentiments, emotions, opinions, appraisals, evaluations, and attitudes towards entities such as services, products, individuals, topics, organizations, and events (Liu 2012). SA has significant importance to organizations, industrial fields, and society, making it one of the most studied research fields (Serrano-Guerrero et al. 2015). It also provides many opportunities to develop new applications, mainly due to the tremendous growth in sources such as social networks. Some examples of these applications are recommender systems, spam detection systems, and question answering systems. Besides, it has been used in many systems for different purposes, such as tourism (Alaei et al. 2019), online education (Kastrati et al. 2020), and transportation (Ali et al. 2019).

The problem of the SA can be identified as follows (Liu 2012): an opinion is considered as a quadruple (**g**, **s**, **h**, **t**), and the SA is the process of finding this quadruple (**g**, **s**, **h**, **t**), where **g**: the destination object for which the sentiment is conveyed, **s**: the sentiment about the destination object, **h**: the person expressing the sentiment (i.e., holder), and **t**: the time of expressing the opinion.

## 3.2 Levels of Sentiment Analysis

Sentiment analysis has usually been studied mainly at three levels: Document level, Sentence level, and Aspect level (Liu 2012); each of these analysis levels is described as follows:

**a. Document Level**

This level is known as document-level sentiment classification that aims to determine whether the whole document expresses an overall opinion (i.e., positive, neutral, or negative) about an item (Pang et al. 2002). For example, given a movie review, the system identified whether the movie's overall opinion is reflected in the review. This level assumes that the whole document discusses only one topic and expresses its opinion (Liu 2012; Schouten&Frasincar 2016). Obviously, this assumption is not applicable for many situations that consist of multiple opinions for multiple entities.

**b. Sentence Level**

This level focuses on analysing a given sentence to determine whether the sentence expresses a positive, negative, or neutral opinion, where neutral indicates no opinion (Schouten&Frasincar 2016). This level comes with a similar assumption to the previous level in that only one sentence can contain sentiment about one topic. Therefore, this level is not applicable in some instances since multiple entities are often compared within the same sentence. This analysis level is closely linked to the subjectivity classification, which differentiates objective sentences from subjective sentences (Ravi&Ravi 2015). Where an objective sentence expresses factual information and a subjective sentence expresses subjective sentiments.

### c. Aspect Level

At both the previous two levels, the calculated sentiment scores are not directly correlated with the entities discussed in the text, where sentiment can be calculated over any specific piece of text (Schouten&Frasincar 2016). In contrast, aspect level performs finer-grained analysis by directly looking at the opinion (i.e., opinion consists of a target and a sentiment) itself, instead of looking at the construction of the language (phrases, sentences, paragraphs, or documents) (Liu 2012; Ravi&Ravi 2015). Therefore, the main essence of this level is to find the sentiment associated with aspects of the entity being discussed in a given text (Serrano-Guerrero et al. 2015). This level was earlier named by Hu and Liu (2004) as feature-based opinion mining and summarization, and it tries to find out what exactly individuals like or do not like. Recognizing the significance of opinion targets allows us to understand the problem of SA better. For instance, while the phrase "although the movie is not that great, I still love the story" obviously has a positive tone. It is difficult to claim that this phrase is wholly positive. The phrase is actually positive about the story but negative about the film. This level is more complex compared to the previous two levels because there are two different types of opinions: regular opinions and comparative opinions (Jindal&Liu 2006). The regular opinion expresses a sentiment only for a particular entity or an aspect of the entity. For example, "Pizza tastes very good," which expresses a positive sentiment on Pizza's aspect taste. While the comparative opinion usually compares multiple entities based on their common aspects, for example, "Pizza tastes better than Donut," which compares Pizza and Donut based on a specific aspect (tastes) and expresses a preference for Pizza. A more detailed description of this level is in section 3.5.

## 3.3 Sentiment Analysis Tasks

There are many tasks related to the SA; some have many aspects in common, thus making it difficult to separate them clearly. The most important ones are (Ravi&Ravi 2015; Serrano-Guerrero et al. 2015): sentiment classification, subjectivity classification, opinion summarization, opinion retrieval, and sarcasm and irony.

### a. Sentiment Classification

Sentiment classification, also referred to as sentiment polarity, opinion orientation, or sentiment orientation, aims to classify the opinions (i.e., sentiments) into one of the three sentiment categories: positive, neutral, or negative (Liu 2012; Serrano-Guerrero et al. 2015). It assumes that document d (e.g., a movie review) reflects the opinions of a single entity e and includes opinions of a single opinion holder h (Liu 2012). This task appears to be a straightforward one, but it is a very complicated one; it is closely linked to the task of sentiment rating prediction. It measures each sentiment's intensity by either predicting the class label in case the rating scale is binary or predicting the rating scores in case it is in a specific range (e.g., 1 to 5) (Serrano-Guerrero et al. 2015).

### b. Subjectivity Classification

Subjectivity classification deals with identifying "private states" - a concept involving opinions, beliefs, evaluations, and emotions (Montoyo et al. 2012). It aims to detect whether a given sentence is subjective or not, where the subjective sentence expresses personal information (e.g., opinions, beliefs, evaluations), not factual information, as discussed earlier. Moreover, the subjective sentence can convey a positive or negative sentiment, but not all of them do. Thus subjectivity classification can guarantee a good sentiment classification if used as a previous step before classification (Serrano-Guerrero et al. 2015).

### c. Opinion Summarization

Mostly, the SA applications need to analyse people's opinions and search for an opinion of a single person is typically inadequate because of the nature of opinions. Opinion summarization focuses on extracting opinions targets and sentiments about them within one or several documents (Serrano-Guerrero et al. 2015). Opinion summarization can be displayed as in one of many forms (e.g., short text summary) of multi-document text summarization. The major elements of a summary should contain opinions about multiple entities, and their aspects should also have a quantitative perspective (Liu 2012). When the summary is based on aspects, it is referred to as an aspect-based (or a feature-based) opinion summary (ABOS) (Hu&Liu 2004). There are two critical features of ABOS, capturing the essential opinion targets (i.e., entities and their aspects) and their sentiments, and it is classified as quantitative. The quantitative side has a high significance due to the nature of opinions; it presented the percentage of people who have a specific opinion (i.e., positive or negative) about each entity and its aspects (Liu 2012).

### d. Opinion Retrieval

The opinion retrieval task is similar to the opinion search that contains two main components: retrieving relevant documents for each query and classifying the retrieved documents (Liu 2012). For the first component, the conventional information retrieval is performed, while for the second component, the retrieved documents are classified as opinionated or not-opinionated. Furthermore, it is required for this task to compute two different scores for each document to rank it; the scores are query relevance score and query opinion score (Serrano-Guerrero et al. 2015).

### e. Sarcasm and Irony

Sarcasm and irony is one of the most complex tasks in SA. It aims to detect sarcastic sentences within a document (Serrano-Guerrero et al. 2015). Sarcasm is a sophisticated form of speech; the speaker writes or says contrary to what they mean. In the SA context, when a user writes something positive, he/she means the opposite (i.e., negative) and vice versa. The reason behind this task's complexity is that there is no agreement among researchers on how sarcasm can be formally described (Filatova 2012). The sarcasm is common with the political discussions, not in the services (i.e., products, hotels, restaurant) reviews (Liu 2012).

## 3.4 Sentiment Analysis Approaches

Sentiment analysis can be roughly divided into two approaches: a machine learning approach and a lexicon-based approach. These approaches aim to determine/classify the user sentiments/opinions for each entity in the given text. Generally, both document-level sentiment classification and sentence-level sentiment classification rely on machine learning approaches. On the other hand, aspect-based sentiment classification depends on the lexicon-based approach (Ravi&Ravi 2015). Figure 9 presents the significant divisions of the SA approaches.

### a. Machine Learning Approach

This approach involves using machine learning for SA; the success of this approach mainly depends on the extraction and selection of the proper set of features used to detect opinions/sentiments (Liu 2012; Serrano-Guerrero et al. 2015). As a result, Natural Language Processing (NLP) techniques play an essential role in this approach. Simply, in this approach, machines automatically learn how to detect user sentiments towards an entity and know whether this sentiment is positive, neutral, or negative. It can be further divided into supervised-learning and unsupervised-learning approaches (Ravi&Ravi 2015; Serrano-Guerrero et al. 2015).

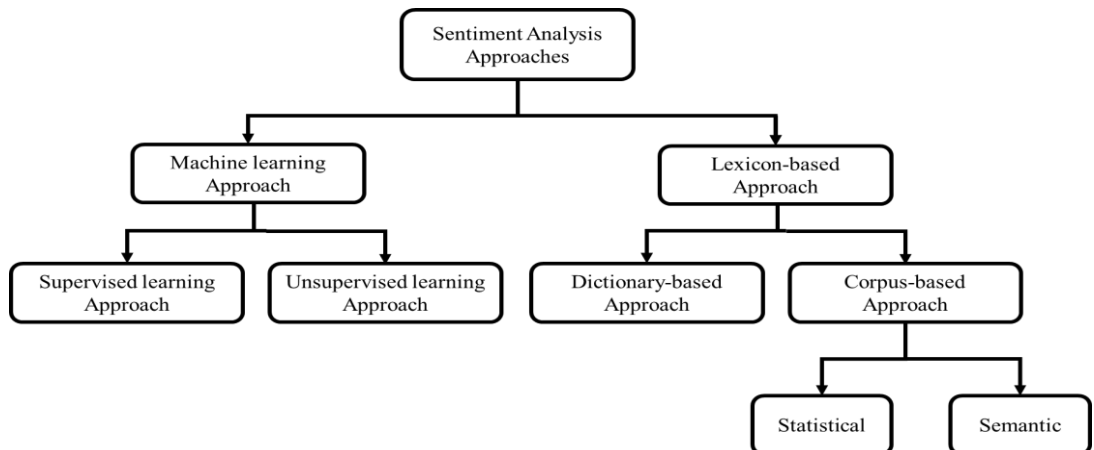

Figure 9      Sentiment analysis approaches

i.   **Supervised Learning Approach**

As earlier discussed, sentiment classification is the primary task for SA, and most researchers consider SA a classification problem. Sentiment classification is described as a two-class classification problem (i.e., positive and negative); most works do not use the neutral class to ease the classification problem (Liu 2012). Basically, sentiment classification is the same problem as text classification; text classification can be defined as a NLP problem aimed to assign one or more predefined categories (topics) for an unclassified document (Al-Ghuribi&Alshomrani 2013). The core feature in text classification is the words related to each topic to ensure the correct classification for the document. In contrast, in the sentiment classification, the critical feature is the opinion words (i.e., words that reflect a positive or negative opinion such as excellent, bad). Since sentiment classification is a problem of text classification, it is possible to apply any existing supervised learning methods (Liu 2012). The most commonly used methods are Support vector machines, Decision trees, Naive Bayes, Maximum entropy, and Neural Networks (Al-Ghuribi&Alshomrani 2013; Hemmatian&Sohrabi 2019; Ye et al. 2009).

ii.  **Unsupervised Learning Approach**

The unsupervised learning approach can be used in the sentiment classification since sentiment/opinion words are usually the primary factor for this classification (Liu 2012). In the supervised learning approach, the annotated/ labelled data are required for training, and the labelling process requires human effort, which is time-consuming and expensive (Hemmatian&Sohrabi 2019). Unsupervised learning addresses this problem. It is proposed when it is difficult to have a set of labelled data and has shown good accuracy in different applications. Clustering approaches are used as unsupervised methods for sentiment classification, and it is first introduced by Liu and Zhang (2012). The clustering approaches can generate reasonably accurate analysis results without any linguistic skills, training time, or human participation (Hemmatian&Sohrabi 2019).

b.   **Lexicon-based Approach**

Sentiment words are words used to express positive or negative opinions. It can also be named opinion words, opinion-bearing words, or polar words, and these words are of great importance for analysis for the SA (Liu 2012). Positive sentiment words express the user's good feeling about a specific entity (e.g., good, fabulous, beautiful). In contrast, the negative sentiment words reflect the user's negative feeling towards an entity (e.g., bad, ugly, awful). A collection of sentiment words are called a sentiment lexicon. Lexicon-based approaches primarily depend on such sentiment lexicon; sentiment lexicon is a collection of precompiled and known sentiment words (Serrano-Guerrero et al. 2015). Lexicon based approaches can be either dictionary-based or corpus-based approaches (Liu 2012).

### i. Dictionary-based Approach

Generally, dictionaries contain a list of words ordered alphabetically and a meaning/synonym for each word. As a result, using the dictionary-based approach to compile sentiment words will be an obvious process (Liu 2012). The sentiment words dictionary is generated by firstly defining a seed list (i.e., a small list of sentiment words) that are manually collected and annotated. After searching in a specific dictionary, this list is then grown by adding the new synonyms and antonyms of the words in the list. The searching is continued until no more new words occur in the searching dictionary (Liu 2012; Ravi&Ravi 2015; Serrano-Guerrero et al. 2015). The previous simple technique for the dictionary-based approach was used by (Hu&Liu 2004; Strapparava&Valitutti 2004). Then, the approach becomes more sophisticated by adding extra processes, such as Kim and Hovy (2004), who attempted to remove the errors from the resulting words and assign a strength for each word in the list using a probabilistic method. The main drawback of this approach is the inability to deal with context-specific orientations and domains (Schouten&Frasincar 2016; Serrano-Guerrero et al. 2015). One of the most popular used dictionaries developed for SA is SentiWordNet, developed by Baccianella et al. (2010), which used the WordNet (Miller 1995) dictionary during its development.

### ii. Corpus-based Approach

The corpus-based approaches arise to solve the dictionary-based approach's drawback by generating dictionaries related to a specific domain (Serrano-Guerrero et al. 2015). This approach depends on the syntactic rules or patterns that come together with an opinion words seed list to find the reminder opinion words within a corpus (i.e., the corpus is an extensive collection of written texts) (Medhat et al. 2014). Two main scenarios have been applied to this approach: either given an opinion words seed list (a general-purpose), and the aim is to find the reminder sentiment words from a domain corpus. Or transform a general-purpose sentiment lexicon to a domain sentiment lexicon using a domain corpus (Liu 2012). This approach can be carried out using two approaches: a statistical approach or a semantic approach (Ravi&Ravi 2015).

For the statistical approach, the aim is to find the co-occurrence opinion words in a corpus using a statistical technique (Medhat et al. 2014). The idea behind this approach is that the polarity/score of a sentiment word can be identified by calculating the frequency of occurrence of this word within a corpus (Read&Carroll 2009; Serrano-Guerrero et al. 2015). In other words, if the word appears more frequently in positive textual content, the polarity of the word is positive, and if it occurs more regularly in negative textual content, its polarity is negative. As a result, if two words often occur together in the same textual content, they are likely to have the same polarity. Thus, by measuring the relative frequency of co-occurrence with another word, an unknown word's polarity can be calculated (Medhat et al. 2014). Using pointwise mutual information (PMI), this could be achieved where PMI measures the degree of statistical dependence between two words. PMI calculates the word's orientation with a set of positive words from its association's strength minus the strength of its association with a group of negative words. Statistical approaches are used in many applications related to the SA, such as the statistical approach Latent Semantic Analysis that analyses the relationship between a group of documents and the words mentioned in these documents (Serrano-Guerrero et al. 2015).

While the semantic approach depends on the similarity between words to directly gives sentiment values for the sentiment words (Ravi&Ravi 2015). In this approach, semantically close words are given similar sentiment values (Medhat et al. 2014). Both Wordnet and SentWordnet dictionaries provide different semantic relations between words that can be used in this approach. This approach is used in many applications that can be used in SA, such as building a lexicon model for describing nouns, verbs, adjectives, and adverbs as the work of (Al-Ghuribi et al. 2020; Maks&Vossen 2012). Moreover, this approach can be integrated with the statistical approach to provide more accurate results.

## 3.5 Aspect-Based Sentiment Analysis

Sentiment analysis has three levels of analysis, as stated before, and the analysis focused on either the document level or the sentence level is not sufficient for many real-life applications. The research method moves closer to opinion targets and opinions on these targets in such applications because the user needs to know specific details such as what entities or their aspects are liked and not liked by others. The previous two levels of analysis still do not achieve that (Liu 2012). In other words, in the previous levels, the calculated sentiment scores are not directly correlated with the entities discussed in the text. For example, a positive opinion towards an entity does not imply a positive opinion about all the aspects of the entity and vice versa (Schouten&Frasincar 2016). Therefore, there is a need to perform finer-grained analysis by directly looking at the opinion (i.e., opinion consists of a target and a sentiment) itself, instead of looking at the construction of the language (phrases, sentences, paragraphs, or documents) (Liu 2012; Ravi&Ravi 2015). Aspect-based sentiment analysis achieves more in-depth (i.e., finer-grained) analysis. It aims to find the sentiment associated with aspects of the entity being discussed in a given text to understand better the SA problem (Serrano-Guerrero et al. 2015).

The Aspect-based sentiment analysis (ABSA) problem can be identified similar to SA as follows (Liu 2012): an opinion is considered as a quintuple ($e_i$, $a_{ij}$, $s_{ijkl}$, $h_k$, $t_l$), and the ABSA is the process of finding this quintuple ($e_i$, $a_{ij}$, $s_{ijkl}$, $h_k$, $t_l$), where: $e_i$ is an entity name, $a_{ij}$ is an aspect j of the entity i, $s_{ijkl}$ is the sentiment score of the aspect j of the entity i, $h_k$ and $t_l$ are the sentiment holder and the time of expressing the opinion respectively. The ABSA problem was first defined by Hu and Liu (2004). Since then, it has increasingly become a trend attracting many researchers' attention since it helps to understand fine-grained opinion mining over coarse-grained (Do et al. 2019). It has a noticeable effect in many applications when the aspects are extracted efficiently with their opinions' scores. For example, they can be used to construct user and item profiles in decision-making processes, such as recommender systems (Da'u et al. 2020; Osman et al. 2019). ABSA focuses on two main tasks: aspect extraction and aspect sentiment classification (Liu 2012; Nazir et al. 2020).

**a.      Aspect Extraction**

Aspect extraction (AE) is one of the NLP tasks aimed to extract the aspects mentioned in text reviews about a particular item (Hu&Liu 2004; Qiu et al. 2011). Aspect usually refers to a concept representing a topic of an item in a specific domain, such as *price, taste, service, and cleanliness,* which are relevant aspects of the *restaurant* domain. Essentially, aspects can be either explicit or implicit (Liu 2012); the aspect is called an explicit aspect when it is directly mentioned in a text review or expressed as a noun or noun phrase. On the other hand, the implicit aspect is the one that is not expressed as a noun or noun phrase and is not directly mentioned in a text review. For example, the following review: "The camera is expensive" contains an implicit aspect 'price' and the opinion word 'expensive' expresses the sentiment of the implicit aspect, whereas in the following review: "The camera has a high price", the aspect 'price' is explicit because it is directly mentioned and accompanied by its sentiment word 'high'. Although some works focused on extracting the implicit aspects, such as (Caputo et al. 2017; Poria et al. 2014; Qiu et al. 2011), most of the notable published works have concentrated on extracting the explicit aspects.

This task was initially studies as a text summarization problem to extract the aspects mentioned in reviews to generate a summary as the work (Hu&Liu 2004; Popescu&Etzioni 2007). Later, two approaches are proposed for achieving this task: supervised and unsupervised. Supervised approaches require labelled aspects, whereas the unsupervised approaches do not require such label datasets for the aspect extraction process.

**b.      Aspect Sentiment Classification**

Aspect sentiment classification (AC) task focuses on identifying the orientation of sentiment/opinion expressed on an aspect with a given text. The critical feature in AC is that an opinion always has a target (i.e., aspect) in an opinion text, so it is important to determine each target's opinion from the text. This task aims to determine the sentiment score $s_{ijkl}$ of the aspect $a_{ij}$, whether it's positive, neutral, or negative (Liu 2012). AC is considered a two-class classification problem since most of the works ignore the neutral texts mentioned earlier; in the datasets with a numeric rating score, the two classes (i.e., positive and negative) should be determined based on the rating values. For example, in the Amazon dataset, the rating scale is from 1-5, so the reviews with ratings ranging from 1 to 2 are considered negative, reviews with ratings equal to 3 are considered neutral, and the reviews with ratings range from 4 to 5 are considered positive. For achieving the AC task, the two approaches: the machine learning approach and the lexicon-based approach discussed in section 3.4, can be used to accomplish the classification task. The majority of the ABSA works rely on the lexicon-based approach to assign a sentiment polarity/score for each opinion word follows by aggregating opinions to determine the final orientation of the sentiment polarity on each aspect in the given text (Liu&Zhang 2012; Ravi&Ravi 2015).

Examples of the used lexicons are SentiWordNet (dictionary-based) and Labille's lexicon built using a corpus-based approach (Labille et al. 2017). A brief description for both of them as follows:

- SentiWordNet, developed by Baccianella et al. (2010), is one of the most successful lexical resources used to determine a sentiment polarity/score for an opinion word. It is derived from the WordNet dictionary, in which every WordNet synset is connected with different three sentiment scores, Pos(s), Neg(s), and Obj(s) ranged from 0.0 to 1.0. These sentiment scores describe how positive, negative, and objective the words contained in the WordNet. Thus, it has around 117,569 sentisynsets (Liu 2012; Medhat et al. 2014; Ravi&Ravi 2015).
- Labille's lexicon is a domain-specific lexicon developed by Labille et al. (2017) using a large corpus. It uses a hybrid method to estimate the sentiment score via the probability method and the information theory method. This hybrid method's efficiency has been proved in the SA process on many large, diverse corpora and overcomes the poor performance of supervised machine techniques.

## 3.6 Aspect-Based Sentiment Analysis Approaches

The main goal of the ABSA is to identify the aspects and compute the sentiment associated with each aspect mentioned in a given text. As discussed earlier, the sentiment score assignment can be done either using: the machine learning approach or the lexicon-based approach. In addition, aspect extraction or aspect detection can be achieved using: supervised machine learning methods or unsupervised machine learning methods (Nazir et al. 2020; Schouten&Frasincar 2016). Figure 12 depicts the different methods for the AE problem.

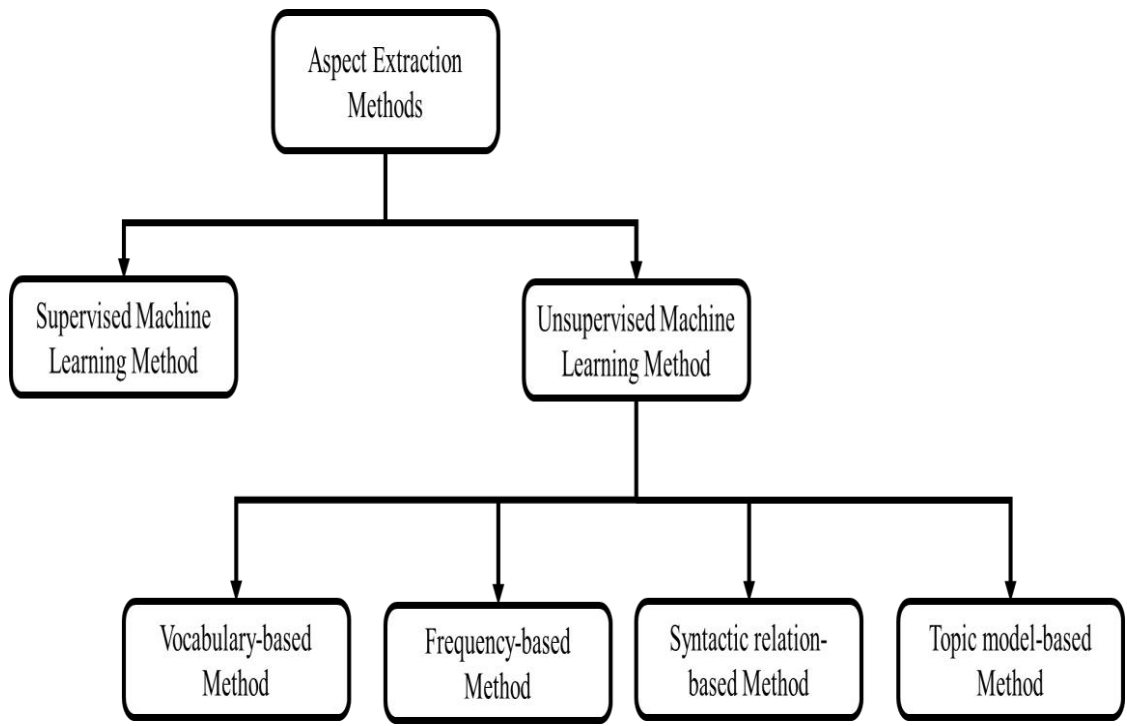

Figure 10    Aspect extraction methods

**a.    Supervised Machine Learning Methods**

The AE problem can be considered as a particular case of the general information extraction problem; thus, many supervised machine learning methods can be used for such a problem (Liu 2012). Essentially, only a few supervised methods that are purely machine learning methods are used for the AE problem since the core key in the supervised method is the extracted features. Feature extraction usually consists of other methods such as frequency-based methods to generate practical features (Schouten&Frasincar 2016). Additionally, it is needed to label data for the AE problem using supervised methods to annotate the aspect words and non-aspect words in the data, which is expensive and time-consuming, especially for large-scale datasets (Liu 2012).

Two supervised methods are mostly used for the AE problem: Conditional Random Fields (CRF) (Lafferty et al. 2001) and Hidden Markov Models (HMM) (Rabiner 1989). The works of (Jakob&Gurevych 2010; Nakagawa et al. 2010) applied the CRF method for the AE problem, while (Jin et al. 2009) applied a lexicalized HMM model for extracting the aspects. Recently deep learning methods have been used for the AE problem in a supervised learning manner, such as the work of (Poria et al. 2016) that used the Convolutional Neural Network method and the work of (Li&Lam 2017) that used the Recurrent Neural Network method.

**b.     Unsupervised Machine Learning Methods**

Most of the current work aimed to solve the AE problem focus on unsupervised methods as supervised learning requires data annotation, which is time-consuming (García-Pablos et al. 2018) and suffers from domain adaptation problems (Darwich et al. 2020). The unsupervised method has been adopted to avoid depending on labelled data since there is no need to separately perform extraction and categorization to obtain the aspects (He et al. 2017). The unsupervised methods for ABSA can be classified into four categories (Al-Ghuribi et al. 2020; Hernández-Rubio et al. 2019): vocabulary-based, frequency-based, syntactic relation-based, and topic model-based methods.

**i.      Vocabulary-based Method**

The vocabulary-based method is the most direct method for AE in which a fixed pre-defined list for aspects (i.e., vocabulary) is used. Few researchers rely only on the pre-defined list for identifying and extracting aspects, such as Aciar et al. (2007), while others use it to extract other aspects related to the elements in the list (Siering et al. 2018). It was claimed that learned aspects (i.e., aspects extracted from the reviews) generate better overall sentiment results than the fixed pre-defined aspects. This is due to the fact that the number of the aspects in the fixed pre-defined list is limited, and there is no guarantee that these aspects will occur in the users' reviews (Caputo et al. 2017). As a result, few researchers use the vocabulary-based method.

**ii.     Frequency-based Method**

The most used method for extracting the learned aspects is the frequency-based method (Hussain&Cambria 2018; Madhoushi et al. 2019). Despite its simplicity, it is very effective and used by many researchers (Hernández-Rubio et al. 2019). The main idea of this method is to extract high occurrences words in reviews (i.e. words that are frequently mentioned by users to express their opinions). The candidate words for aspects are the noun and noun phrase (Caputo et al. 2017; Eirinaki et al. 2012; Hu&Liu 2004; Moghaddam&Ester 2010; Mowlaei et al. 2020; Mubarok et al. 2017). If the candidate word's frequency exceeds some threshold value, the word is considered an aspect. Once the aspect is extracted, the aspect's sentiment word is selected based on the nearest adjective to the aspect. Finally, the selected sentiment word is assigned a polarity value (i.e. a score) based on some lexicons. Both Caputo et al. (2017) and Mubarok et al. (2017) are among the recent works that use the frequency-based method. Follows are the details of their works.

Caputo et al. (2017) proposed a system for opinion retrieval called a sentiment aspect based retrieval engine (SABRE), which consists of four tasks: extract aspects and their sub-aspects; find the opinion associated with each aspect; detect the polarity for each sentiment opinion word and retrieve documents for a given opinion. The extraction of aspects is based on term-frequency probabilities and a model that calculates the difference distribution for a word between a specific domain and a general corpus using the non-symmetric Kullback-Leibler divergence technique. For opinion word extraction and its polarity, a lexicon-based approach is applied using the AFINN wordlist. Finally, the TF-IDF weighting scheme retrieves the top-N documents with the highest opinion scores for document retrieval. The experiment of the proposed method on the TripAdvisor dataset containing 167,780 reviews outperformed the conventional term frequency method in terms of F-measure.

On the other hand, Mubarok et al. (2017) proposed an approach for extracting sentiment polarity based on specific aspects of product reviews. The model creates two bag-of-words models where one of the models contains the aspects (i.e., nouns), and the other includes the sentiment words (i.e., adjectives or adverbs). The words in both lists are selected using a chi-square test by choosing the words with the highest relevance for each opinion. The Naive Bayes classifier is then used to classify the sentiment polarity of each aspect. The model is evaluated using the SemEval-2014 dataset, which contains 3,618 reviews for the restaurant domain consisting of five aspects (i.e., price, food, ambience, service, and miscellaneous). The model is compared against 17 baselines. The proposed method received the seventh-highest F-measure among the 17 compared baselines with an F-measure = 78.12%.

Although the frequency-based method is an efficient one, it has obvious limitations. One of the limitations is that the approach may select words that are not aspects (i.e., pick up many words that do not contain any subjectivity) because it relies only on word frequencies (Schouten&Frasincar 2016). Furthermore, aspects that are not frequently mentioned will not be detected using this method. However, the syntactic relation-based and topic model-based methods can address such a limitation.

### iii. Syntactic Relation-based Method

The syntactic relation-based method (also called the rule-based method) analyses the sentence's syntactic structure and the relations among the words to identify the aspect's sentiment words. A well-known algorithm that uses this method is the Double Propagation proposed by Qiu et al. (2011). The algorithm describes the syntactic relation between nouns or noun phrases with adjectives using dependency grammar. This method has been used as a baseline for many other methods, while others, such as Poria et al. (2014), tried to improve it by expanding the relations extraction rules. Further research use dependency parsing to define the relationships among the words (Chen&Yao 2010; Dragoni et al. 2019; Hai et al. 2013; Luo et al. 2019; Mukherjee&Bhattacharyya 2012; Nejad et al. 2020). Following are the description of some notable works based on this method as follows:

Chen and Yao (2010) presented an approach defining the opinion words' relations using both dependency parsing and shallow semantic analysis, then built an ontology and a collocation (i.e., the most frequently co-existing topic and sentiment pairs') database. The method is applied to two datasets used in (Hu&Liu 2004; Lakkaraju et al. 2011), containing 500 and 2500 reviews. Both datasets are mainly from product domains such as laptops, cameras, printers, and DVDs. The proposed method outperforms two baselines, a naive baseline in which the final polarity is calculated by the major number of positive or negative opinion words in the sentence, and (Hu&Liu 2004) as baseline 2, in terms of accuracy.

Mukherjee and Bhattacharyya (2012) proposed a method for identifying the features and their corresponding opinions in the product reviews using dependency parsing to define the short and long dependencies between words. An experiment is performed using a Chinese corpus on the car domain collected from Internet product reviews. The proposed approach results outperform the compared baselines (of closest-pair and dependency parsing) in terms of precision, recall, and F-measure. Hai et al. (Hai et al. 2013) used three syntactic dependency rules with two domain corpus to extract the aspects from two domains (i.e., cellphones and hotels) of Chinese reviews. The candidate aspects are first extracted using the three rules. These aspects are filtered based on their relevance to the domain using two relevance measures: intrinsic-domain relevance and extrinsic-domain relevance. The F-measure results of their conducted experiments on the Chinese reviews were 63.6% and 52.2% for the two domains, respectively.

Nejad et al. (2020) is one of the recent research that employed an unsupervised approach for detecting explicit features in the Persian language for the hotel domain. Their methodology consists of three steps, text preprocessing, sentimental vocabulary construction, and aspect extraction. A directed weighted graph is constructed based on frequent pattern identification from the sentences of their Persian corpus. The paths within the constructed graph are determined based on some developed rules to extract multi-word aspects. The proposed approach is evaluated and compared with some existing approaches that work on the Persian language, and it gives the best F-measure value.

Like the frequency-based method, the syntactic relation-based method produces noise in terms of non-related aspects due to only focusing on the subjective expressions (i.e., sentence structures) and ignoring the words' semantics. Besides, not all the used rules are effective for extraction and not all the extraction pattern rules are explored (Tubishat et al. 2021).

### iv. Topic Model-based Method

The topic model-based method addresses the syntactic relation-based method problem by focusing on the semantics of the words. This is because the method reveals topics from an extensive collection of texts whereby words are grouped into aspects or topics. For example, users talk about price using words like money, budget, and cost, which should not be regarded as different aspects. The topic modelling is based on two basic models, Latent Dirichlet Allocation (LDA) (Blei et al. 2003) and Probabilistic Latent Semantic Analysis (Hofmann 2001), for learning latent topics (i.e., the local topics) that have a direct correlation with the aspects (i.e., the general topic). It is used by many research (Brody&Elhadad 2010; Cheng et al. 2018; García-Pablos et al. 2018; Lin&He 2009; Moghaddam&Ester 2011; Seroussi et al. 2011; Wang et al. 2010). For example, in Brody and Elhadad (2010) work, a local LDA is applied to the restaurant reviews dataset by assuming that each sentence in the reviews is a separate document to extract low frequent aspects. The works of Lin and He (2009) and Moghaddam and Ester (2011) extend the standard LDA.

In the work of Lin and He (2009), an additional sentiment layer is added to the basic LDA model to develop a Joint Sentiment Topic Model, in which the topic word is not separated from the sentiment word. The work of (Moghaddam&Ester 2011) extends the LDA to Interdependent Latent Dirichlet Allocation based on the assumption that there is an interdependency relation between the aspect and the sentiment words. In other work, McAuley et al. (2012) proposed a probabilistic model that benefited from the ratings associated with the reviews to learn words that correlate with aspects or specific ratings. For instance, the word appearance may be used to represent the look aspect, and the word delicious may refer to a high rating. To build this model, three learning methods are used: supervised, semi-supervised, and unsupervised. They introduced a new dataset consisting of 5,000,000 reviews, where each user provides ratings for each aspect of a product. The model is evaluated on three prediction tasks: determining the review's parts that discuss the rated aspects, finding the sentences that determine the user's rating, and predicting the non-rating user's aspects. The authors claimed that their model is suitable for real datasets and can determine the reviews' parts related to each aspect and select the sentence that best summarizes the review.

In recent years, deep learning approaches generally received particular attention in SA studies and in ABSA, particularly Zhou et al. (2019). LDA is combined with deep learning techniques to enhance the aspect extraction process (Alqaryouti et al. 2019; Chauhan et al. 2020; Ekinci&İlhan Omurca 2020; García-Pablos et al. 2018; Kumar et al. 2020; Liang et al. 2020). The topic modelling approach with deep learning is illustrated by referring to the works of Garcia et al. (2018) and Chauhan et al. (2020).

Garcia et al. (2018) proposed an unsupervised approach called W2VLDA, which is based on topic modelling combined with continuous word embeddings and a maximum entropy classifier. The approach consists of three subtasks: aspect classification, sentiment classification, and aspect/opinion word separation. The approach's performance is evaluated in the multilingual SemEval-2016 task 5 dataset (Pontiki et al. 2016). It is tested for three domains, electronic devices, restaurants, and hotels and for four languages, English, French, Spanish and Dutch. Therefore, it outperforms two of the conventional LDA-based approaches. Chauhan et al. (2020) integrated the rule-based method with Bidirectional Long-Short-Term-Memory (Bi-LSTM) model to extract the aspects. The rule-based method is used to extract the candidate aspects from nouns and noun phrases, followed by the Bi-LSTM model to filter the candidate aspects and select the correct ones. Like Garcia et al. (2018), the SemEval-2016 dataset is used to evaluate the restaurant and laptop domains. The approach's results missed many aspects during the extraction process because it only considers nouns and noun phrases (Tubishat et al. 2021).

Some recent works combine both Conditional Random Field (CRF) and (Bi-LSTM) models in the aspect extraction process, such as Liang et al. (2020) and Gandhi & Attar (2020). The former use SemEval 2014 and 2015 (Pontiki et al. 2015) datasets, and the latter use Hindi dataset to evaluate their approaches. While in other work, an extension of LSTM is proposed, such as Ma et al. (2018). They proposed two methods; the first is a Sentic LSTM that contains a separate output gate that interpolates both concept-level input and token-level memory. The second method is an extension of the Sentic LSTM; it merges the LSTM and a recurrent additive network that simulates sentic patterns. The performance of the proposed methods is evaluated on both SentiHood dataset and SemEval 2015 dataset. Results show the effectiveness of the proposed methods in both aspect categorization and aspect-based sentiment classification tasks.

The topic model-based method has two main limitations (Ngoc et al. 2019; Schouten&Frasincar 2016). First, it's a restriction of use in real-life SA applications because the method will not achieve reasonable and efficient results if the data size is small. This makes such a method unsuitable for many practical SA applications. The second limitation is that the result of LDA contains more global topics than local ones because it is designed for the document level, thus using it at the aspect-level is not a wise decision.

## 4 CONCLUSION

Currently, user-generated reviews are used to improve the accuracy of the RSs performance by using sentiment analysis to transform the unstructured user reviews into a structured form that can be merged with RSs. This paper provides a thorough overview of the recommender system and sentiment analysis. To begin, it provides an overview of the recommender system, covering phases, techniques, and performance indicators. The paper then goes over the sentiment analysis concept and covers the most important points of sentiment analysis, such as level, approaches, and aspect-based sentiment analysis. We expect this survey will help researchers to gain more understanding about recommender system and sentiment analysis.